# Unsupervised Abnormality Detection Using Heterogeneous Autonomous Systems

Sayeed Shafayet Chowdhury
Department of ECE
Purdue University
email: chowdh23@purdue.edu

Kazi Mejbaul Islam
Department of EEE
Ahsanullah University of Science and Technology
email: kazimejbaul@ieee.org

Rouhan Noor
Department of EEE
Ahsanullah University of Science and Technology
email: rnlio1995@gmail.com

*Abstract*— Anomaly detection (AD) in a surveillance scenario is an emerging and challenging field of research. For autonomous vehicles like drones or cars, it is immensely important to distinguish between normal and abnormal states in real-time. Additionally, we also need to detect any device malfunction. But the nature and degree of abnormality may vary depending upon the actual environment and adversary. As a result, it is impractical to model all cases a-priori and use supervised methods to classify. Also, an autonomous vehicle provides various data types like images and other analog or digital sensor data, all of which can be useful in anomaly detection if leveraged fruitfully. To that effect, in this paper, a heterogeneous system is proposed which estimates the degree of abnormality of an unmanned surveillance drone, analyzing real-time image and IMU (Inertial Measurement Unit) sensor data in an unsupervised manner. Here, we have demonstrated a Convolutional Neural Network (CNN) architecture, named AngleNet to estimate the angle between a normal image and another image under consideration, which provides us with a measure of anomaly of the device. Moreover, the IMU data are used in autoencoder to predict abnormality. Finally, the results from these two algorithms are ensembled to estimate the final degree of abnormality. The proposed method performs satisfactorily on the IEEE SP Cup-2020 dataset with an accuracy of 97.3%. Additionally, we have also tested this approach on an in-house dataset to validate its robustness.

## I. Introduction

The autonomous and intelligent vehicle is one of the promises of the fourth industrial revolution of machine intelligence, blockchain and the Internet of things. Detecting abnormalities of the autonomous vehicle has become a popular research field as it's important for providing security and ensuring stability of the vehicle by learning and detecting abnormalities of the device by gathering sensory data [1]. Also determining normal/abnormal dynamics in a given scene from external viewpoint is one of the emerging research fields [2-3]. This paper proposes an approach for detecting device abnormalities by using intelligent and heterogeneous systems in an unsupervised manner where we determine abnormalities of a drone that is being used for surveillance.

Nowadays, different methods based on machine learning coupled with signal processing are used to develop various novel methods and models by research in areas like biosensing [4-6], image classification [7], assistive technologies [8-12], and heart rate measurement [13-14]. Campo *et al.* [15] proposed a method to detect abnormal motions in real vehicle situations based on trajectory data where they used Gaussian process (GP) regression that facilitates the whole vehicle's movement over sparse data. Kanapram *et al.* [16] proposed a novel method to detect abnormalities based on internal cross-correlation parameters of the vehicle with Dynamic Bayesian Network (DBN) to determine the abnormal behavior. Iqbal *et al.* [17] proposed a method where they selected an appropriate network size for detecting abnormalities in multisensory data coming from a semiautonomous vehicle. As previous works mostly rely on a high level of supervision to learn private layer (PL) self-awareness models [18-24], Ravanbakhsh *et al.* [25] proposed a dynamic incremental self-awareness (SA) model which allows experiences done by hierarchical manner, starting from a simpler situation to structured one. In this paper, means of cross-modal Generative Adversarial Network (GAN) is used for processing high dimensional visual data. Baydoun *et al.* [26] also proposed a method based on multi-sensor anomaly detection for moving cognitive agents using both external and private first-person visual observations to characterize agents' motion in a given environment where the semi-unsupervised way of training as a set of Generative Adversarial Networks (GANs) was used that produce an estimation of external and internal parameters of moving agents. Most of the works focus on environmental anomaly detection, however the device anomaly is not taken into consideration. If the device is anomalous itself, then even if the anomaly detection system is working well, the whole pipeline will be hampered greatly. Keeping this in mind, the IEEE SP CUP 2020 competition [27] focuses on autonomous device anomaly detection in a surveillance setting, which is the problem addressed here.

In this paper, we introduce a Siamese Network [28] like regression block which consists of a Convolutional Neural Network (CNN) architecture namely AngleNet to estimate angle displacement between two images. In addition, we have used autoencoder based anomaly detection system for Inertial Measurement Unit (IMU) data and ensembled all those outputs in such a manner that can estimate the degree of abnormality of the device for given image and IMU data samples at a timestamp. The main contributions of this paper are:

1. To detect device anomaly, we have used a modified Siamese-like regression network to estimate angle displacement between two images in a regression manner.

2. We have ensembled the anomaly of angle with the outputs of two autoencoders which are used to estimate degree of anomaly of two separate IMU sensor data and calculated overall anomaly of the drone.

The rest of the paper is organized as follows. Section II explains the details of the problem description. Section III describes the proposed method where the AngleNet and the rest of the clustering methods are explained in detail. Section

IV presents the experimental result and comparison and finally, Section V describes the conclusion of the work.

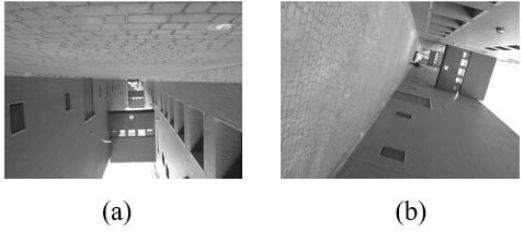

Figure 1: (a) and (b) are examples of normal and abnormal images respectively

## II. PROBLEM AND DATASET DESCRIPTION

In SP Cup 2020 [27], data of an unmanned drone was provided which contained data from IMU sensor and images of respective time frames. Rosbag files are provided which contains compressed sensor data and images. Some files contained only normal time frames while other files contained both normal and abnormal time frames which are mixed. The task was to find the abnormal time frames using unsupervised methods which means we had to use only normal data for training and other calculations and using it we had to find the abnormal cases. Total 12 Rosbag files where total number of normal images is 277 and the number of mixed images is 392.

Besides image data, they have provided IMU sensor data. There are 6 types of data under IMU topicname among which we have used IMU/data and IMU/mag. A total of 987 normal timestamps was provided. IMU/data contains the orientation and velocity information of the drone along 3 axes and IMU/mag contains magnetic field data read by magnetometer. There are two separate parts in this detection procedure, image analysis and IMU sensor data anyalysis. We used autoencoder based anomaly detection system for IMU data and used AngleNet to estimate angle of input image with respect to a normal image sample and later ensembled the 3 outputs to estimate degree of abnormality. Figure 1 shows some normal and abnormal image samples provided in this dataset and it is very clear that angle of view is changed significantly between the two cases.

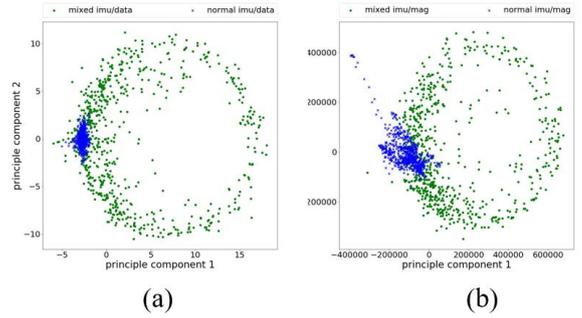

Figure 2: PCA plot of (a) IMU/data and (b) IMU/mag

To detect anomaly, we need to differentiate between normal and abnormal data. Since they are inherently high-dimensional, we first explore them in a lower dimensional sub-space using PCA analysis which is shown in Figure 2. From the figure, it is evident that normal and abnormal data form separate clusters in the 2D space spanned by the first two principal component directions. This indicates that there is inherent differentiable characteristics in the normal and abnormal type of data, which we try to distinguish more vividly and accurately in the next section using our method.

## III. PROPOSED METHOD

This section describes the method we have used to develop an unsupervised model to detect abnormalities using the image and IMU sensor data. It is very clear that when the abnormal images were taken, the drone due to some anomaly became rotated at a significant angle. To accurately estimate the angle without depending on the abnormal data, we have introduced AngleNet which is used in an unsupervised manner. Besides we have used autoencoder based algorithm for modeling normal IMU data. Figure 3 demostrates how we have combined the three outputs altogether to estimate degree of anomaly. As per the figure, rather than using the outputs of the models directly, we have integrated weights manually and estimated final degree of anomaly. The intuition is, AngleNet can only estimate the angle, not measure it precisely and for a single timestamp, the rotation of the drone may be slightly higher than the threshold angle but the sensor data may be significantly anomalous. Also it can happen vice versa. So it is appropriate to use weights and these are discussed further in Section IV. The different blocks of the proposed approach are described next in details.

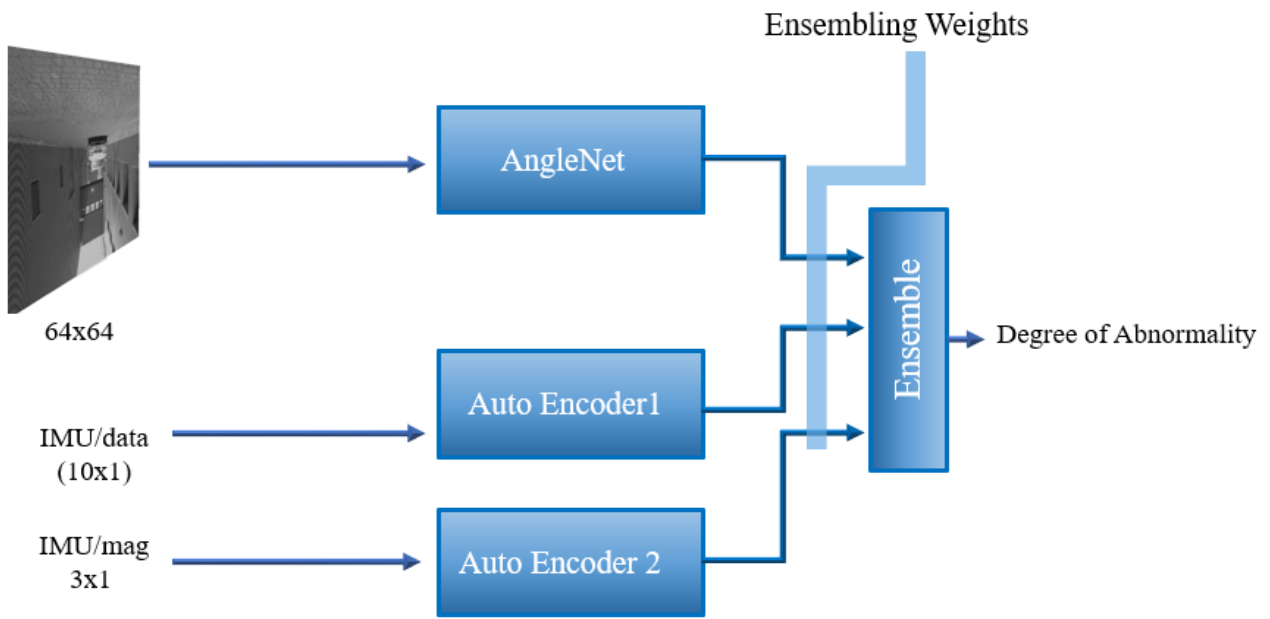

Figure 3: Flow chart for estimating degree of abnormality

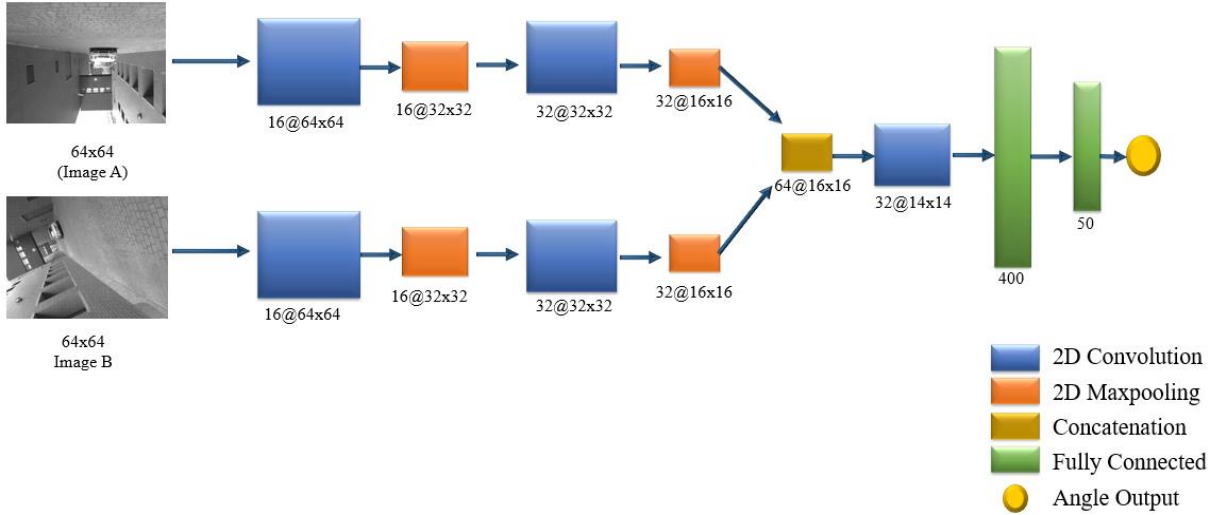

Figure 4: AngleNet used for estimation of angle difference between two images

### A. AngleNet

In the abnormal state of a surveillance drone, it is mostly the tilt angle that varies from the normal state as depicted in Fig. 2. In normal conditions, the drone is pretty stable as shown in the dataset. While for the unstable drone, the image is tilted at a significant angle. Inspired by SIAMESE network [28] we introduce AngleNet, a convolutional neural network architecture to estimate significant angle change from the normal state. Previously Spyros *et al.* [29] introduced RotNet but we can not use this model in this case as it works in a classification manner and can onlye detect angles among 0, 90, 180, 270 degrees. But in AngleNet we tried to demonstrated ange estimation as a regression problem which is reasonable. In this model, a normal image should be provided first, and then the upcoming frames will be taken as input and the output is the angle between them. If there is a significant difference between the images such as object mismatch, the output will be significantly high which is the reason why we can not use [30] or any other classical computer vision based angle estimation systems. In Figure 4 we see the model structure where it takes two images at size of 64x64 pixels and. On the layers, 16@64x64 means a tensor having a depth of 16 channels and height and width of 64 rows and columns respectively.

As discussed in section II, there are not much image data taken by drone to train AngleNet in supervised manner. To use this model in an unsupervised manner in this case, we have used Stanford car dataset [31] to pre-train. The images were first augmented as so it can mimic the angle change, as demonstrated in Figure 5. The activation function of the final layer was relu as it is a non-negative linear function. Finally there were 48,000 images which was divided into 80% train image and 20% validation image. Mean Squared Error was used as the loss function.

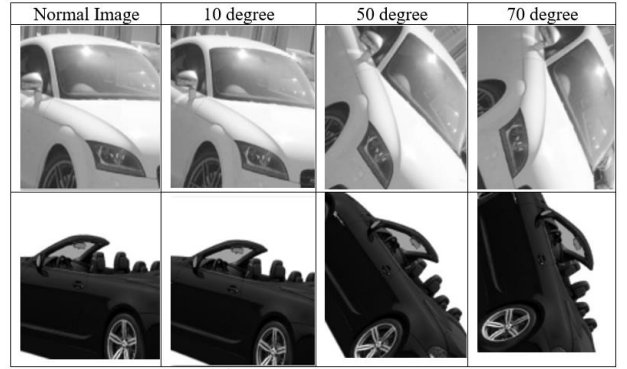

Figure 5: Samples of normal images and augmented images from Stanford Car Dataset [31]

After pre-training the model using the dataset, we have fine-tuned using the provided dataset [27] to estimate the degree of abnormality $\sigma_d$ of the images by dividing the output angle by 90.0 degree. The performance of the model on classifying normal images is demonstrated in Section IV.

### B. Autoencoder based AD of IMU data

For both the IMU/data and IMU/mag, we have used autoencoder based anomaly detection system, similar to [32] but not exactly. The autoencoder models used here are shown in Fig. 6. While training the autoencoders, it is supposed that in abnormal time frame, both IMU/mag and IMU/data reading are abnormal. As there is no clear description of the dataset on this issue, we have considered as such and it produced good result in our practical experimentation which is described in Section IV-F.

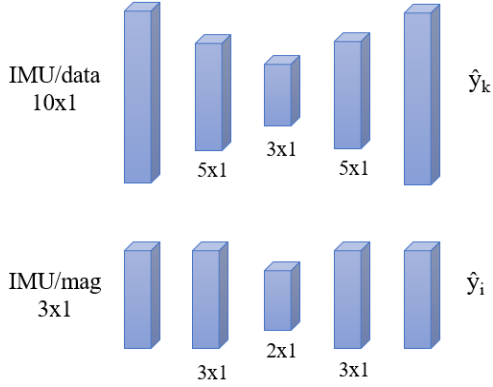

Figure 6: Autoencoder architecture used for IMU/data and IMU/mag

As we are considering both are abnormal or normal at the same time frame, we have trained two autoencoders together. In this case, mean squared error(mse) was considered as reconstruction loss. If L1 is the mse for IMU/data sample and L2 is mse for IMU/mag sample, then:

$$L_1 = \frac{1}{n} \sum_{k=0}^{n}(y_k - \hat{y}_k)^2 \quad (1)$$

$$L_2 = \frac{1}{n} \sum_{i=0}^{n}(y_i - \hat{y}_i)^2 \quad (2)$$

Where, $y_k$ and $\hat{y}_k$ are ground truth and reconstructed output respectively for IMU/data samples. And $y_i$ and $\hat{y}_i$ are ground truth and reconstructed output for IMU/mag samples. And training loss is defined by the linear addition of the two losses, L1+L2. While training these two autoencoders, we have used normal samples only as we have discussed earlier why it is not good practice to use supervised classification for normal and abnormal samples. The results are discussed in Section IV.

To find degree of anomaly in this manner we had to consider $L_{max,\,data}$ and $L_{max,\,mag}$ as maximum reconstruction errors from the reconstruction errors of all individual samples from IMU/data and IMU/mag respectively. And calculated $\sigma_d$, $\sigma_m$ using (3) and (4).

$$\sigma_d = \frac{L_1}{L_{max,data}} \quad (3)$$

$$\sigma_m = \frac{L_1}{L_{max,mag}} \quad (4)$$

## IV. EXPERIMENT AND RESULT

In this section, we have discussed the training procedures and results on SP Cup dataset [27]. As this dataset is novel and contains two types of data, we have discussed the results separately and lastly we have ensembled the results and produced final output.

### A. IMU Anomaly Detection

As discussed in Section II, we have used autoencoder based anoaly detection system for IMU data. The results are discussed in Table 1. Normalization improves performance for IMU/mag as it has a wide range of data, from (-400000, 400000), it is very tough to converge the loss without normalization. The authors in [33] have claimed better accuracy than us but their system is resource expensive.

TABLE 1. PERFORMANCE ON IMU DATA

| Data | Accuracy | F-1 Score | False Negative |
|---|---|---|---|
| IMU/data | 97.8% | 0.9812 | 2 |
| IMU/mag | 100% | 0.95 | 0 |

We are using two tiny autoencoders with shared loss for IMU anomaly detection which can be run on common embedded devices of low resource like raspberry pi, which is validated in subsection *G*. Even though their accuracy excels us, we can run our system in much lower resource having very few or no false negative cases. If there is reported a device anomaly falsely, it is okay in this case as it will work like a false alarm with no harm for very few cases. The comparison among some other popular algorithsm are given in Table 2. Our proposed approach was among the top 8 performing ones in IEEE SP Cup 2020 using clustering for IMU analysis, we later improved it using autoencoder.

TBALE 2: COMPARISONS WITH OTHER ALGORITHMS

| Algorithm | Accuracy |
|---|---|
| Autoencoder, Rad *et. al.* [34] | 96.87% |
| Kmeans Clustering, Iqbal *et. al.* [17] | 93.28% |
| 1D CNN, Kiranyaz *et. al.* [35] | 89.9% |
| PCA and Kmeans | 82.3% |
| Spectral Clustering | 91.7% |

### B. AngleNet Based Anomaly Detection

Using AngleNet, we can estimate the angle between the test image and the normal image. In the abnormal images, the rotation angle is the main distinctive factor. Any images rotated by 30 degrees is supposed to be abnormal. But the threshold is perfectly tunable and user-defined. The performance of AngleNet on the test images is shown in Table 3.

TABLE 3. PERFORMANCE OF ANGLENET ON NORMAL IMAGE

| Threshold Angle | Accuracy |
|---|---|
| 30 | 94.7% |
| 20 | 86.4% |

Table 3 shows the change of accuracy level by changing the threshold angle. The state of the art image/video novelty detection algorithms are mostly for environmental anomaly detection which is not perfectly inclined with the problem we have worked with. We are more interested to find the anomaly of device rather than the environment. Some comparisons with various methods are showon in Table 4.

TABLE 4: COMPARISON OF ANOMALY DETECTION USING IMAGE

| Algorithm | Accuracy |
|---|---|
| AngleNet | **94.7%** |
| Optical flow supervised | 89.4% |
| Optical flow, unsupervise | 91.33% |
| Binary Classification | 84.85% |
| Autoencoder | 91.7% |

Angle is not only the difference between normal and abnormal image samples, also there are some motion factor. So we have compared the performance of anomaly detection between using optical flow and actual images. In the dataset, the provided images were sampled so they did not have a gradual movement shift among them, rather there is a rapid difference in motion and content, so optical flow did not show a good performance. Some examples of Dual TVL1 optical flow of normal and abnormal images are shown in Table 5. Here the right most sample is from normal time stamp and the others are from abnroal timestamps. The image from Figure 2 (a) was used as reference normal image to calculate all the optical flows.

TABLE 5: SAMPLES OF OPTICAL FLOW

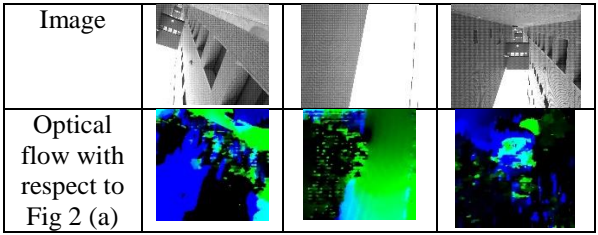

| Image | | | |
| Optical flow with respect to Fig 2 (a) | | | |

## C. Ensembling Models

The process is designed so that both the clustering-based anomaly detection and AngleNet can be used separately or in an ensemble manner. As we have discussed we calculate the degree of abnormality in each case, ensembling them can produce a combined result. Proposed ensembling formula is given as-

$$N = w_d * \sigma_d + w_m * \sigma_m + w_I * \sigma_I \quad (5)$$

Where N is the combined degree of abnormality, $w_d$, $w_m$, $w_I$ represents the weights for three different models such as two autoencoder-based models and one Convolutional Neural Network-based model, AngleNet. And $\sigma_d$, $\sigma_m$, $\sigma_I$ represent the degree of abnormality for IMU/data, IMU/mag, and Image respectively. After ensembling those models, we have achieved overall 97.8% accuracy with F-1 score of 0.98. To find the 3 weights, first we considered 1 for each of them and performed extensive experiments to empirically determine the values of the weights. Our investigation finds $w_I$ = 0.75, $w_m$ = 0.9 and $w_d$ = 1 provides the best results and these values were chosen for subsequent analysis. We considered abnormal if N ≥ 1.

## D. Computational Cost

While training AngleNet on the Car dataset, we have used GoogleColab with Nvidia Tesla K80 GPU with 12 Gigabytes of memory. But for the testing purpose, it runs on a computer with 2 Gigabytes of GPU seamlessly. The system is tested on a system containing the Intel Core i5 processor, 8 Gigabytes of RAM and Nvidia 940 MX. It takes 0.47 seconds on an average to process a single frame.

## E. Implementation on In-house setup

For demonstrating real-time usage on an embedded device, it is used on a raspberry pi where the autoencoders ran on the pi and CNN based processing works on a remote server. The system is tested on in-house setup, with custom hexacopter running on Ardupilot and used raspberry pi 3 for real-time processing and sending video frames to the server. In this setup, the accuracy of the algorithm was 96.49%.

## V. CONCLUSION

In this paper, we have demonstrated an ensembled approach for unmanned vehicle anomaly detection for surveillance purposes. Our approach does not classify any sample as strictly normal or abnormal, rather we have used the degree of abnormality. The lower this value of anomaly, the closer it is to normal situation. We have introduced AngleNet which is used to estimate angles between two input images and using this angle we estimate degree of anomaly. We have trained AngleNet on Stanford Card Dataset and used tranfer leanring to estimate angle of the images of IEEE SP Cup 2020 dataset. For detecting abnormal IMU samples, we have used autoencoder based anomaly detection system. Besides maintaining good accuracy, our proposed method has the advantage of being low-cost and it is possible to execute anomaly detection in real-time. Future work could be to integrate defense mechanism from adversarial attacks which is a potential threat to any surveillance system.